\documentclass[lettersize,journal]{IEEEtran}
\usepackage{amsmath,amsfonts}
\usepackage{algorithmic}
\usepackage{algorithm}
\usepackage{array}
\usepackage[caption=false,font=normalsize,labelfont=sf,textfont=sf]{subfig}
\usepackage{textcomp}
\usepackage{stfloats}
\usepackage{url}
\usepackage{verbatim}
\usepackage{graphicx}
\usepackage[backend=biber,sorting=none]{biblatex}
\begin{document}

\title{VTP: Volumetric Transformer for Multi-view Multi-person 3D Pose Estimation}

\author{Yuxing Chen, Renshu Gu\IEEEauthorrefmark{1}, Ouhan Huang, Gangyong Jia\IEEEauthorrefmark{1}
        % <-this % stops a space
\thanks{Manuscript received xxx. The work was supported by the the National Key
Research and Development Program under Grant No. 2019YFC0118404, National Natural Science Foundation of China under Grant No.U20A20386, Zhejiang Provincial Science and Technology Program in China under Grant No. LQ22F020026, Fundamental Research Funds for the Provincial Universities of Zhejiang under Grant No.GK219909299001-028, Zhejiang Key Research and Development Program under Grant No. 2020C01050, the Key Laboratory fund general project under Grant No. 6142110190406, the key open project of 32 CETC under Grant No. 22060207026 (Corresponding authors: Renshu Gu; Gangyong Jia.)}
\thanks{The authors are with the Hangzhou Dianzi University, China. (email: chenyuxing@hdu.edu.cn, renshugu@hdu.edu.cn, huangouhan@hdu.edu.cn, gangyong@hdu.edu.cn)}% <-this % stops a space
}



\maketitle

\begin{abstract}
This paper presents Volumetric Transformer Pose estimator (VTP), the first 3D volumetric transformer framework for multi-view multi-person 3D human pose estimation. VTP aggregates features from 2D keypoints in all camera views and directly learns the spatial relationships in the 3D voxel space in an end-to-end fashion. The aggregated 3D features are passed through 3D convolutions before being flattened into sequential embeddings and fed into a transformer. A residual structure is designed to further improve the performance. In addition, the sparse Sinkhorn attention is empowered to reduce the memory cost, which is a major bottleneck for volumetric representations, while also achieving excellent performance. The output of the transformer is again concatenated with 3D convolutional features by a residual design. The proposed VTP framework integrates the high performance of the transformer with volumetric representations, which can be used as a good alternative to the convolutional backbones. Experiments on the Shelf, Campus and CMU Panoptic benchmarks show promising results in terms of both Mean Per Joint Position Error (MPJPE) and Percentage of Correctly estimated Parts (PCP). Our code will be available.
\end{abstract}


\section{Introduction}
\IEEEPARstart{M}{ulti-view} multi-person 3D pose estimation \cite{zhang2021direct} has been an essential task that benefits many
real-world applications, such as surveillance, intelligent sports, virtual/augmented reality, and human-computer interaction. It aims to localize the 3D joints for all people from multi-view cameras. Compared to single-view approaches, multi-view cameras can provide complementary information to effectively alleviate projective ambiguities.

One of the biggest challenges in multi-view multi-person 3D pose estimation is that the identity of the persons from each camera view is unknown. Previous methods \cite{dong2019fast,huang2020end} tried to tackle this problem as an association task. First, the same persons from different views are identiﬁed and associated across multiple views. Subsequently, each person's pose is estimated by triangulation or optimization-based
pictorial structure models. Establishing cross-view correspondences is often crucial for this line of research. Recently, some approaches \cite{tu2020voxelpose,wu2021graph} bypass the cross-view matching problem and perform both human localization and pose estimation in the 3D volumetric space. VoxelPose \cite{tu2020voxelpose} proposes to use a 3D object detection design that avoids making incorrect decisions in each camera view, and enables a collaborative decision from all camera views. Nevertheless, while the 3D Convolutional Network (CNN) in VoxelPose is good at extracting local features, CNN has limitations when capturing long-range dependencies. Additionally, the volumetric representation calls for heavy memories and computations, which is an unnecessary overhead for sparse scenes. This paper attempt to explore better representation learning for voxel-based methods in multi-view multi-person 3D pose estimation.

Transformer has demonstrated promising performance in many vision tasks. In this paper, we propose a transformer framework that learns volumetric representations for multi-view multi-person 3D pose estimation. Inheriting the advantages of VoxelPose \cite{tu2020voxelpose}, VTP aggregates features from 2D keypoints in all camera views and directly learns the spatial relationships in the 3D voxel space. The aggregated 3D features are passed through 3D convolutions before being flattened into sequential embeddings and fed into a transformer. A residual structure is designed to further improve the network performance. 

A major bottleneck of transformer-based models lies in that the self-attention computation is quadratic and the computational cost is unbearable if applied directly to voxels. Towards this end, VTP overcomes this bottleneck from two aspects: (1) VTP adopts a coarse-to-fine routine, where persons are first localized in 3D and transformers are applied to the 3D region of interest (ROI) to regress the 3D poses in the second stage; (2) VTP exploits the sparse Sinkhorn attention which computes quasi-global attention with only local windows based on differentiable sorting of the representations, thus improving the memory efﬁciency. 

Our contributions can be summarized as follows:
\vspace{0.25cm}
\begin{itemize}

\item We present the first transformer framework that learns volumetric representations for multi-view multi-person 3D pose estimation, named VTP, which empowers the transformer to perform pose estimation in the 3D volumetric space instead of explicit cross-view matching in 2D.

\item We design a residual block structure with 3D convolutions for VTP to further improve accuracy and performance. 

\item To overcome heavy computations of voxels, we exploit the sparse Sinkhorn attention which computes quasi-global attention with only local windows based on differentiable sorting.

% \vspace{1cm}
\item Extensive experiments on benchmark datasets show that our approach is on par with existing state-of-the-art methods in terms of Percentage of Correctly estimated Parts (PCP), and shows superior Mean Per Joint Position Error (MPJPE) performance compared to previous volumetric representation based methods.

\end{itemize}

\section{Related Work}
\subsection{Multi-view 3D Pose Estimation}
To exploit information from different viewpoints, many methods have been proposed for multi-view human pose estimation
\cite{zhang2021direct,tu2020voxelpose,iskakov2019learnable,qiu2019cross,zhang20204d}. Establishing cross-view correspondences is key to many existing works \cite{iskakov2019learnable,qiu2019cross,zhang20204d}. Iskakov et al. \cite{iskakov2019learnable} propose to fuse 3D voxel feature
maps and learn 3D pose via differentiable triangulation.
Qiu et al. \cite{qiu2019cross} introduce a cross-view fusion scheme into CNN to jointly estimate 2D poses for multiple views, and present a recursive Pictorial Structure Model to recover the 3D pose. Zhang et al. \cite{zhang20204d} propose a deep-learning-free 4D association method that unifies per-view parsing, cross-view matching, and temporal tracking into a single optimization framework. In 2020, Tu et al. propose a volumetric paradigm VoxelPose \cite{tu2020voxelpose} and bypass the cross-view matching problem, thus effectively reducing the impact of incorrectly established cross-view correspondences. Inspired by VoxelPose, we present the first transformer framework for 3D volumetric representation for multi-view 3D pose estimation. MvP \cite{zhang2021direct} views multi-person 3D pose estimation as a direct regression problem and does not perform any intermediate task including 2D keypoint detection. Yet, we argue that the intermediate 2D keypoint detection makes methods more robust to environmental change. 

\subsection{Vision Transformer}

\par Transformer, first introduced in Attention is All You Need \cite{vaswani2017attention}, is a common model in the field of natural language processing (NLP) and has swept various tasks of NLP with its outstanding performance. Thanks to its known advantages in capturing long-range dependencies, the computer vision (CV) field has also recently seen a soaring number of papers using transformers in various tasks, such as object detection \cite{carion2020end, zhu2020deformable, mao2021voxel,guan2022m3detr}, image classification \cite{dosovitskiy2020image,liu2021swin, li2022bvit}, video understanding \cite{sun2019videobert, girdhar2019video, lee2020parameter, wang2021end}, and Super-Resolution \cite{yang2020learning, qiu2021nested, cai2021freqnet}. There is also some recent research on the use of transformers in human pose estimation tasks such as 2D pose detection \cite{mao2022poseur,mao2021tfpose,li2021pose,stoffl2021end}. PRTR \cite{li2021pose} proposes a top-down regression-based method of human pose estimation network based on cascade transformer structure. POET \cite{stoffl2021end} proposes the first end-to-end trainable multi-instance 2D pose estimation model, which combines CNN and transformer to directly predict the pose in parallel using the relationship between the detected person and the whole image environment. 

\par For the 3D pose estimation, METRO \cite{lin2021end} presents a method for reconstructing 3D human pose and mesh vertices from a single image. The attention encoder is used to jointly model vertex-vertex and vertex-joint interactions and to output 3D joint coordinates and mesh vertices simultaneously.
% TransFusion \cite{ma2021transfusion} introduces a transformer framework for multi-view 3D pose estimation, aiming at directly improving individual 2D predictors by integrating information from different views instead of post-prediction merge/calibration. 
However, no one has tried to exploit transformers for volumetric representations in multi-view 3D pose estimation, due to its high computational cost, especially for large sparse scenes. In fact, MvP \cite{zhang2021direct} designs a multi-view pose transformer that is free of volumetric representations. 
\par In contrast, to explore further potentials of the volumetric representations and benefit from its elegant and simple formation, we propose the first volumetric transformer pose estimator (VTP) for multi-view 3D pose estimation. In this paper, we prove that the transformer framework for 3D volumetric representations shows outstanding performance compared to other voxel-based methods.

\begin{figure*}[h]
  \includegraphics[width=\linewidth]{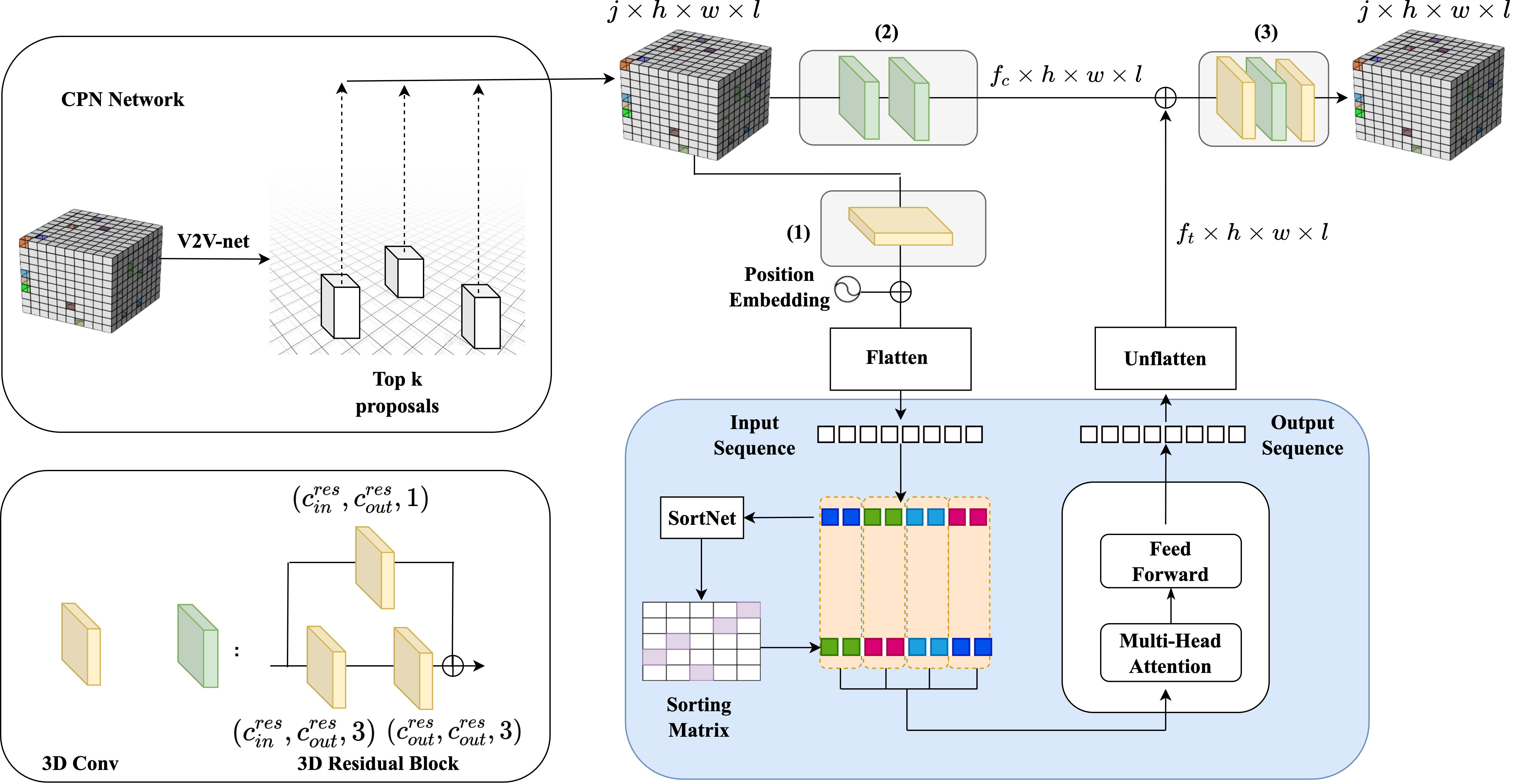}
  \caption{Overview of the network structure.}
  \label{overview}
\end{figure*}

\section{Method}
Figure \ref{overview} is an overview of the proposed framework.  The detection process is a top-down, two-stage approach. First, The 2D heatmaps at each viewpoint are obtained by  pretrained 2D pose backbone. Second, the entire public space will be voxelized, using the Cuboid Proposal Network (CPN) network, the same as VoxelPose\cite{tu2020voxelpose}, to predict each persons' center in the space. Third, the space around each predicted human center will be voxelized in a more detailed space, and the 
VTP network is used to predict all the 3D joint positions of that person.
VTP takes the CPN predicted human center's surrounding voxel grids as input, containing the multi-view fused information about a person’s surrounding space. The input will go through a separate convolution-based network(marked as (2) in Figure\ref{overview}) and a transformer-based network(including (1) in Figure\ref{overview} and following transformer encoder), finally concatenating their outputs and putting them through the convolutional layer (marked as (3) in Figure\ref{overview}).
\par Specifically, given a voxel space $V\in \mathbb{R}^{j\times h\times w\times l}$, where $h$, $w$, $l$ are the height, width, and length of certain grid space surrounding the center of the target person, $j$ denotes the number of predicting joints per person which is predefined in different datasets. In a single grid, there are j numbers, each representing the $j$-th joint’s average possibility predicted from multi-view 2D heatmaps. These grids are fed into the transformer $t\left(\cdot\right)$ and convolutional layer $c\left(\cdot\right)$, which will both extract the $j$ channels into high-level features $X_t=t\left(V\right)\in \mathbb{R}^{f_t\times h\times w\times l},\ X_c=c\left(V\right)\in \mathbb{R}^{f_c\times h\times w\times l}$.  These two outputs are concatenated together and fed into the 3D convolutional network to convert the high-level features into voxel grids $V_o\in \mathbb{R}^{j\times h\times w\times l}$ which have the same channels as the initial input. Finally, the coordinates of each joint node can be obtained by weighting and summing the coordinates of the voxel grids with a special trick\cite{sun2018integral}
\subsection{Voxel Transformer Pose (VTP) Estimator}

%  The detection process is a top-down, two-stage approach. First, the entire public space will be voxelized, using a CPN network to predict each persons' center in the space, following \cite{tu2020voxelpose}. Second, the space around each predicted human center will be voxelized in a more detailed space, using the voxel Transpose network to predict all the joint nodes of that person.

\textbf{Input of the network}. The target people are first localized by a number of  Cuboid Proposal Network proposals, following VoxelPose \cite{tu2020voxelpose}. The 2D pose heatmaps in all camera views are projected to a common discretized 3D space to construct a 3D feature volume. The space is divided into discrete grids of $X\ \times Y\ \times\ Z$, expressed as $\left\{\textbf{V}^{x,y,z}\right\}$. Denote the 2d heatmap under the viewpoint $v$ as $M_v\in\mathbb{R}^{J\times H\times W}$, where $J$ is the number of joints. For each $\textbf{V}^{x,y,z}$, the projection position $P_v^{x,y,z}$ under viewpoint $v$ can be calculated by the camera geometry, then the value of the 2d heatmap at this projection position can be expressed as $\textbf{M}_v^{x,y,z}$. At this point, for a $\textbf{V}^{x,y,z}$, the average value of the 2d heatmaps can be calculated for all the views, $\textbf{F}^{x,y,z}=\frac{1}{V}\sum_{v=1}^{V}\textbf{M}_v^{x,y,z}$, $V$ being the number of views that can observe this space. At this point, every feature vector of $\left\{V^{x,y,z}\right\}$ can be obtained.

\textbf{3D Conv and Residual block}.
There are three places where the 3D convolution is used in this network: (1) the transformer's embedding, (2) the initial input voxel grids are extracted into high-level features,  and (3)   after the output of two branches are concatenated together, the high-level features voxel grids are recovered to original channels. The 3D Conv used here has four parameters: \[c_{in},c_{out},kernelSize,padding\]

\par $c_{in},c_{out}$ represent the input voxel channels and the target output channels, $Kernel Size$ is the size of the 3D convolution kernel,
here we make $padding=(kernelSize-1)/2$. The main function is to transform the channels of each voxel element while keeping the dimensions of the voxel grids constant.

\par  Using three 3D convolutions will form a residual block, where $c_{in}^{res},c_{out}^{res}$ are the two parameters of the residual block. The input of the residual block will go through two branches, one consisting of $c_{in}=c_{in}^{res},c_{out}=c_{out}^{res}$ and $c_{in}=c_{out}^{res},c_{out}$ $=c_{out}^{res}$, two 3D convolution layer with kernel size set to 3. And the other one consists of one 3D convolutional layer with $c_{in}=c_{in}^{res},c_{out}=c_{out}^{res}$ and kernel size of 1. The two paths will result in four-dimensional vectors of the same scale, which are summed and then passed through the ReLU activation function to obtain the final output of the residual block. 
% The residual network is characterized by its ease of optimization and its ability to improve accuracy by adding considerable depth. Its internal residual block uses jump connections, which alleviate the gradient disappearance problem associated with increasing depth in deep neural networks.

% \par All the 3D convolutional blocks adopt the structure of CNN + Batch Normalization + ReLU. Using batch normalization can increase the training speed, prevent overfitting, improve the generalization ability and avoid the activation function from entering the nonlinear saturation region, which causes the gradient dispersion problem.

 \textbf{Position Embedding}. Positional embeddings are there to give a transformer knowledge about the position of the input vectors. They are added (or concatenated) to corresponding input vectors. The learnable positional embedding, same as in Bert\cite{devlin2018bert}, is used in the proposed framework which lets the network derive the effective position embedding in the training data by itself. The authors also tried Sinusoidal Position Encoding in three dimensions: length, width, and height of voxel grids to make the transformer perceive the position of the current grid in 3D space. However, the experimental results were not satisfactory.

\begin{figure}[h]
  \centering
  \includegraphics[width=\linewidth]{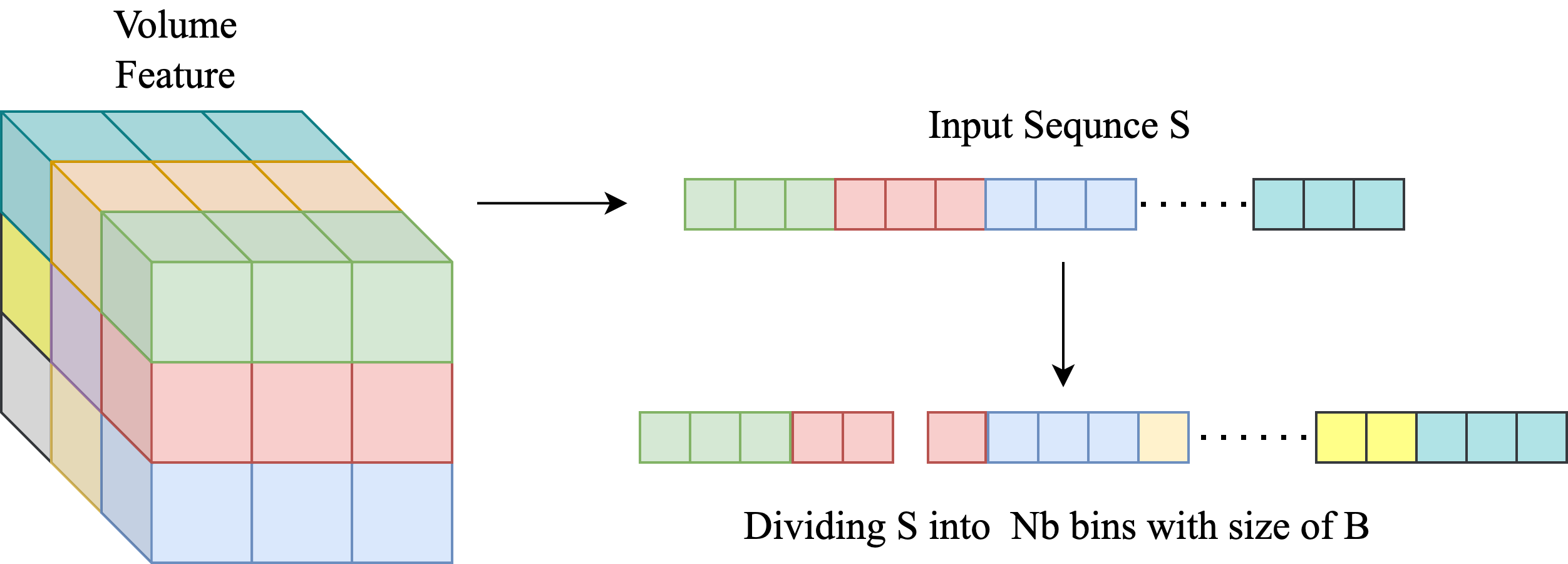}
  \caption{Flatten and dividing process}
  \label{dividing}
\end{figure}

 \textbf{Sinkhorn Sparse Transformer on voxel grids}.
The transformer module $t\left(\cdot\right)$ accepts sequential input and the process of converting four-dimensional voxel grids into two-dimensional sequences is described below. First, let a $h\times w\times l$ 3D grid with $j$ dimensional vectors be $\textbf{V}=\left\{v^{0,0,0},\ldots,v^{h,w,l}\right\}\in\mathbb{R}^{j\times h\times w\times l}$, where $v^{x,y,z}$ represents a grid vector with coordinates $\left(x,y,z\right)$, the dimension $j$ is equal to the number of human joint nodes defined in the dataset. Second, feeding $\textbf{V}$ into a 3D convolutional network with $c_{in}=j,c_{out}=e$ yields a voxel grid with more features extracted as $\textbf{V}_e=\left\{v_e^{0,0,0},\ldots,v_e^{h,w,l}\right\}, v_e\in\mathbb{R}^{e\times h\times w\times l}$, then each $v_e$ will be added to the position embedding of the same length $e$. This completes the embedding process of the initial input. Third, the voxel grids with four dimensions $h,w,l,e$ are transformed into a two-dimensional sequence $\textbf{S} = \left\{s_0,\ldots\ s_L\right\}, s_i\in\mathbb{R}^{e \times L}$, where the mapping relation  is $s_{x+\left(h\times y\right)+\left(h\times w\times z\right)}=v_e^{x,y,z}$,  as shown in Figure \ref{dividing}.

\par For an input sequence $\textbf{S}$ of length $L$, it will be divided into $N_b$ bins in the way of one bin for every $B$ elements. Let the partition function be $\theta\left(\cdot\right)$, $Bins=\theta\left(\textbf{S}\right)$ , where Bins represents all the blocks that are partitioned out, $Bins=\left\{b_1,\ldots,b_{N_b}\right\},b_i\in\mathbb{R}^{B,e}$. 
\par If each element can only focus on other elements in the same bucket as itself, this approach is equivalent to a block-based local attention mechanism that lacks macroscopic dependence on other elements far away from itself, whereas for 3D pose estimation, it is important to infer the likelihood of a joint in the current position based on cues from other distant joints. The following section describes how the Sinkhorn sparse transformer reorders the bins so that the current region can focus on distant regions related to itself, making it an approximate global attention process.
\par The self-attention mechanism in transformers relies on the calculation of the Query, Key, and Value of each element, which can be written as:

\begin{equation}
\resizebox{0.9\linewidth}{!}{

$Attention\left ( Q, K, V \right ) = Score\left ( Q, K \right ) V = softmax
\left ( \frac{QK^T}{\sqrt{d} }  \right ) V $

}
\end{equation}

\par The relevance of element $j$ to element $i$ can be obtained by calculating the dot product of the Query of element $i$ and the Key of element $j$. After the weights are calculated for all other elements, the Value values are summed up according to their relevance. This requires the computation process to store the $QK^T$ matrix quadratic to the length of the input sequence, and here is where the bottleneck of the original transformer lies. For voxel grids of human pose estimation, the grid density is positively correlated with the accuracy, and if the original transformer is used then the grid density will be very sparse and cannot achieve valuable prediction accuracy. Nevertheless, voxel grids and transformers are both sparse, and most of the space in voxel grids does not have human joint nodes. The elements that have a large impact on the final result of the transformer are often very few, so if we can make each element focus only on the most relevant elements with itself, we can reduce the quadratic complexity. The main idea of Sinkhorn sparse attention is to divide the sequence into blocks, calculate the correlation between blocks and learn the appropriate block sorting order. After sorting, each block in the original order will only focus on the elements in the corresponding block in the new sequence. This alleviates the problem caused by the quadratic complexity of transformers and makes it possible to exploit transformers for volumetric representations.

\par  First, each element in each bin $b_i$ is multiplied by three different linear transformation matrices $W_Q,W_K,W_V$ respectively to obtain three new matrices $b_i^Q,b_i^K,b_i^V$, which represent the set of Query, Key, and Value of each element in $i$-th bin. Next, the mean of Query and Key $b_i^{Q_{mean}},b_i^{K_{mean}}$ for each bin is calculated.

\begin{equation}
b_i^{Q_{mean}}\left[i \right]=\frac{1}{B}\sum_{j=1}^{B}{b_i^Q\left[j\right]\left[i\right]},i\in\left[1,...,e\right]
\end{equation}

\begin{equation}
b_i^{K_{mean}}\left[i \right]=\frac{1}{B}\sum_{j=1}^{B}{b_i^K\left[j\right]\left[i\right]},i\in\left[1,...,e\right]
\end{equation}

\par Here the obtained $b_i^{Q_{mean}},b_i^{K_{mean}}$ for dot product can be used to estimate the degree of correlation between two bins, approximating the element-to-element relationship in the original transformer. At this time, the complexity of computing $Score\left (b_i^{Q_{mean}},b_i^{K_{mean}}  \right ) $ for all bins can be reduced from the quadratic of the input sequence length $L$ to $N_b^2$, and the attention matrix is obtained  as follows, where $a_{i,j}$ denotes the overall correlation between bin $i$ and bin $j$:

\begin{equation}
R = \begin{bmatrix}
 a_{0,0} & \cdots & a_{0,N_b}  \\
 \vdots & \ddots    & \vdots \\
 a_{N_b,0}  & \cdots  &  a_{N_b,N_b}
\end{bmatrix}
\end{equation}

\par If the matrix R is doubly stochastic (matrix elements are all non-negative and all row sums and column sums are one), it becomes a permutation matrix. Permutation matrices, which reorder the elements of a vector, are special cases of doubly stochastic matrices. Each permutation matrix is a convex combination of doubly stochastic matrices, so learning doubly stochastic matrices can be considered a form of relaxed permutation matrices. Here a method is needed to compute an arrangement based on the already-computed bin-to-bin correlation matrix, i.e. to compute a permutation matrix with which to multiply the input sequence of bins for sorting.

\par An optimal assignment could be obtained using the Hungarian algorithm. However, this operation of selecting the optimal assignment is not differentiable. In this case, since deep learning models rely on derivative gradient descent for parameter updating, we would not be able to use neural networks for this problem. Can we use a differentiable operation to approximate the operation of picking the optimal assignment so that it can be learned? The answer is yes, and the solution is to use the iterative Sinkhorn normalization.

\par For a matrix $R$ the normalization is performed for rows and columns repeatedly and independently, which is called the Sinkhorn normalization process \cite{adams2011ranking}. $k$
represents the number of user-defined iterations. This process is expressed in the following equations \cite{tay2020sparse}:

\begin{equation}
S^0=exp\left(R\right)
\end{equation}

\begin{equation}
S^k=F_c\left(F_r\left(S^{k-1}\left(R\right)\right)\right)
\end{equation}

\begin{equation}
S\left(R\right)=\lim_{k \to \infty} {S^k}\left(R\right)
\end{equation}
where $F_r,F_c$ denote the row and column regularization functions with the following formula:

\begin{equation}
F_c^k\left(X\right)=F_c^{k-1}\left(X\right)-log\left({exp\left(X1_l\right)\textbf{1}}_N^\top\right)
\end{equation}

\begin{equation}
F_r^k\left(X\right)=F_r^{k-1}\left(X\right)-log\left({\textbf{1}_l\textbf{1}}_N^\top e x p\left(X\right)\right)
\end{equation}

\par Sinkhorn \cite{sinkhorn1964relationship} proves that repeating the above two steps eventually converges to a doubly stochastic matrix if $R$ has support. Each of these steps is derivable so that the chain of derivations required to update the parameters for deep learning is not interrupted by the computation of the permutation matrix, thus enabling end-to-end training.

\par
\begin{figure}[h]
  \centering
  \includegraphics[width=\linewidth]{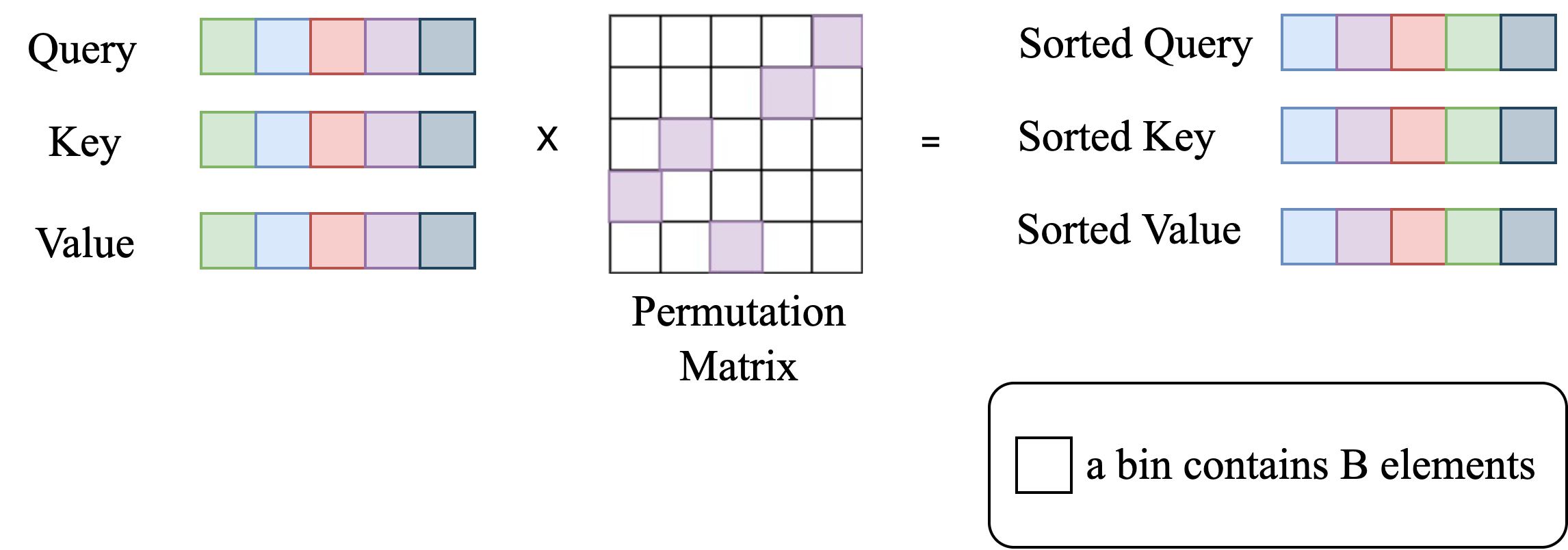}
  \caption{Reordering process.}
  \label{fig:reord_proc}
\end{figure}

 The matrix obtained after the Sinkhorn operation will be used to reorder the initial bin sequence $b_i^Q,b_i^K,b_i^V,i\in\left(1,N_b\right)$ into a new sequence $b_i^{Q_{sorted}},b_i^{K_{sorted}},b_i^{V_{sorted}},i\in\left(1,N_b\right)$, as shown in Figure \ref{fig:reord_proc}.

\par Subsequently, the Query, Key, and Value of the two sequences are concatenated into a bigger bin, which results in a Query, Key, and Value with $2B$ elements. Here the bigger bin contains both locally adjacent elements and other relevant elements from afar.

\begin{figure}[h]
  \centering
  \includegraphics[width=\linewidth]{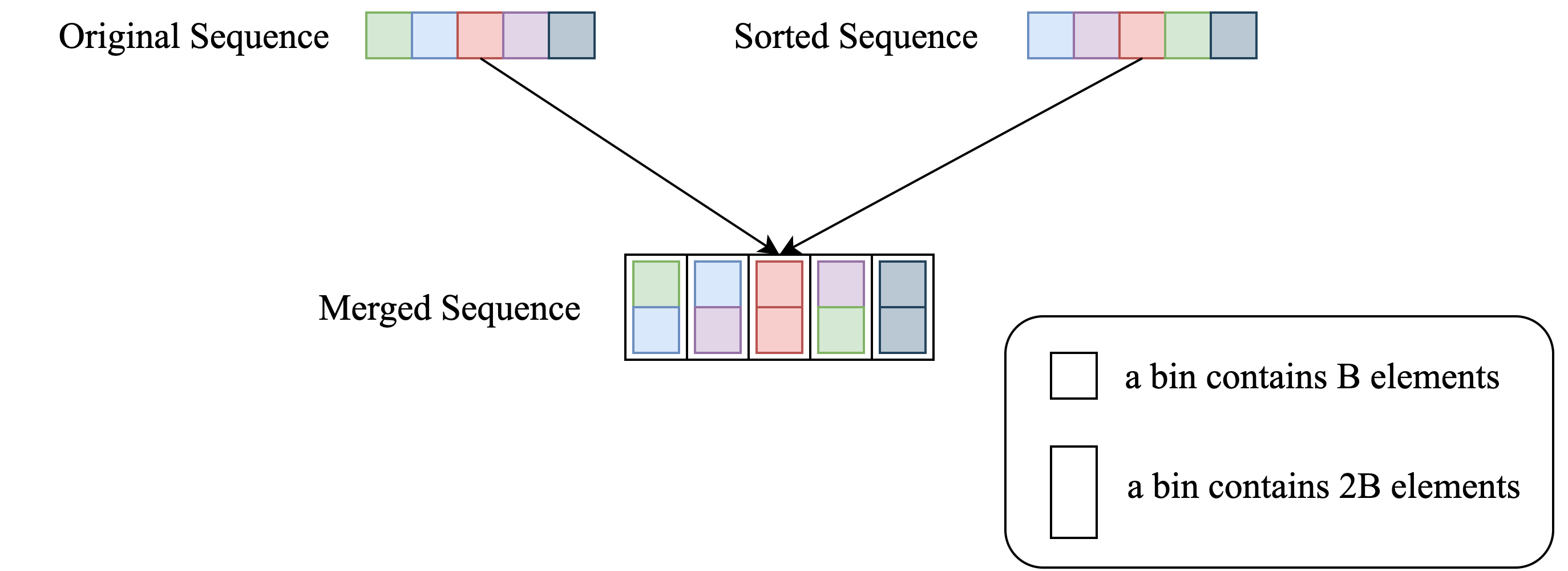}
  \caption{Combining two sequences.}
\end{figure}

% \par The attention is computed using a multi-headed attention mechanism with parallel computation. If there are currently $N$ attention heads, the Attention module will split the query, key, and value parameters into $N$ parts, send them to each of the N attention heads, perform the Scaled Dot-Product attention operation, and finally concatenate the results. Multiple representation ``subspaces" are provided so that the model can pay attention to information from different "representation subspaces" at different locations. Intuitively, multiple heads enable us to attend independently to (parts of) the sequence, analogous to the use of multiple filters in CNN. In this case, the $2B$ elements in each bin are subject to the same operations as in the original transformer, so that the global attention effect in the original transformer can be approximated.

\par Figure \ref{fig:attention_matrix} represents the attention matrix of sparse Sinkhorn transformer. The orange area represents the relevant area after the original sequence is chunked, and the blue area represents the elements in the corresponding position of the sequence after the permutation matrix transformation:

\begin{figure}[h]
  \centering
  \includegraphics[width=0.5\linewidth]{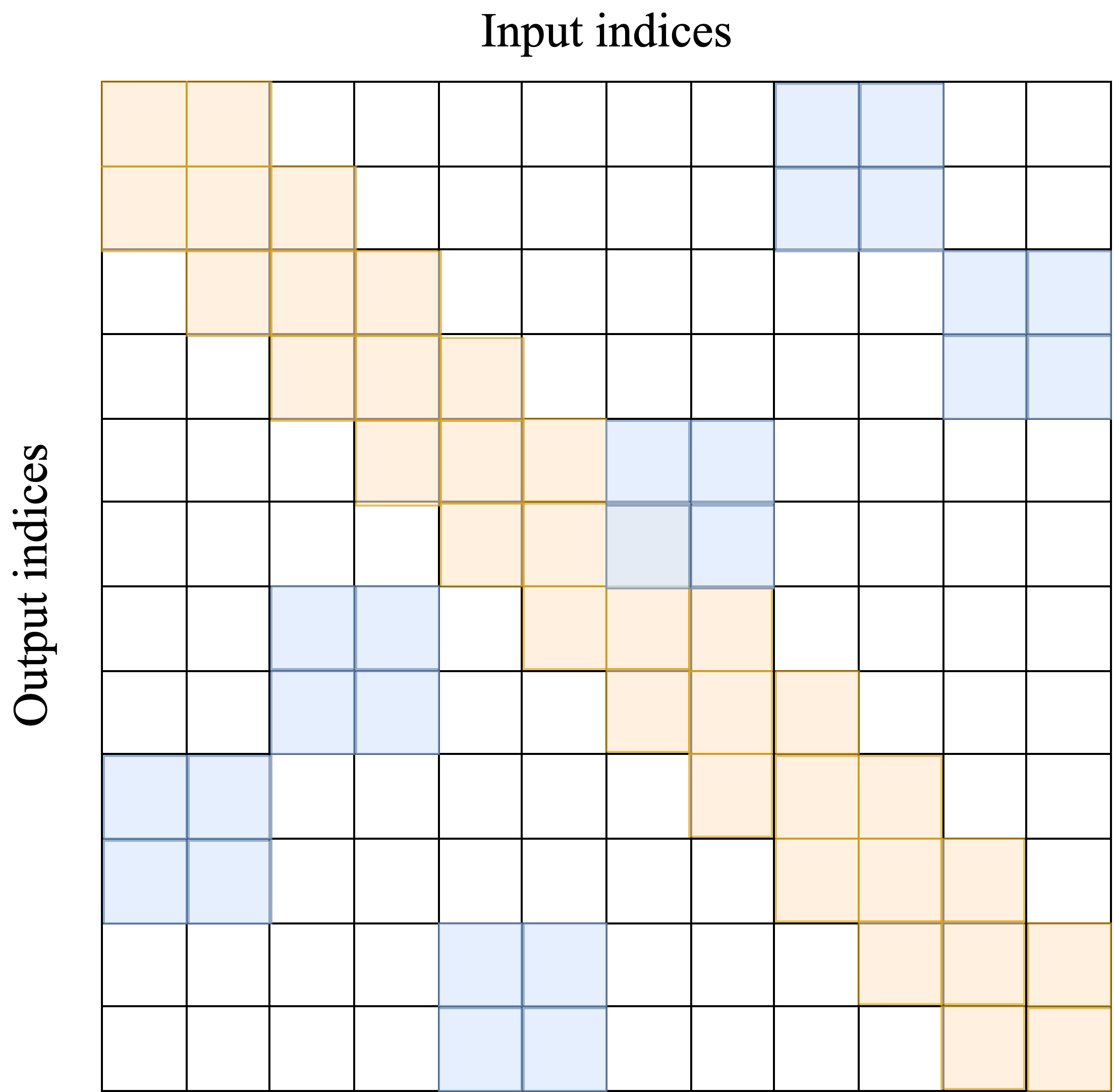}
  \caption{Attention matrix \cite{fournier2021practical}}
  \label{fig:attention_matrix}
\end{figure}

\par The whole process requires computing the attention matrix twice, once for a sequence in bins, with a complexity of $O\left(N_b^2\right)$; the second time for a bin containing $2B$ elements, with a complexity of $O\left(B^2\right)$. So this attentional approach is changed from the original
complexity of $O\left(L^2\right)$ to $O\left(B^2+N_b^2\right)$, where $B=\frac{L}{N_b} $.

% \par The transformer can contain both encoder and decoder, which is suitable for generative tasks such as those that want to generate information different from the input sequence, machine translation, or article summary summarization for instance. Other models such as BERT use the encoder-only structure, which is suitable for tasks such as predictive classification or word embedding. The encoder-only structure is also used in our framework, which mainly extracts deeper information from voxel grids.

\par After the sparse Sinkhorn transformer encoder, the output $S_o\ =\ \left\{s_0,\ldots\ s_L\right\}$ will be the same channels as the input, each element being a vector of $e$ channels. the sequence is unflattened into 3D space again and concatenated with the output from the 3D convolutional network. At this point, the channels of each voxel element will become $e + c$. Finally, the features are fed into the last layer of 3D convolutional network, where input channels equals $e+c$ and output channels is $j$, where $j$ is the number of human joints. It will change the high-dimensional feature into the probability of the existence of each joint in the current voxel space.

\par\textbf{Final output.}
After the above steps, voxel grids with joints possibility will be obtained. However, this is not enough to deduce the final position of each joint node, because if we just pick the most probable value from the used grids, then the $2000mm \times 2000mm \times 2000mm$ space will be divided into $32 \times 32 \times 32$  grids, and each grid will have the length of 62.5mm, the accuracy would be limited here. Similar to \cite{tu2020voxelpose}, we don't find the voxel grid which has the maximum probability as prediction, instead we calculate the weighted average from all voxel grids, which is a common technique in previous works \cite{sun2018integral}.

\par Let $\textbf{P}_j\left(x,y,z\right)$ represent the probability map that the $j$-th joint node is in the voxel grid $\textbf{V}^{x,y,z}$, and then the final position of this joint can be derived from the following equation:

\begin{equation}
\textbf{J}_j = \sum_{x=1}^{X} \sum_{y=1}^{Y} \sum_{z=1}^{Z}\left ( x,y,z \right )\cdot \textbf{P}_j\left ( x,y,z \right ) 
\end{equation}

\par In this way, the prediction probabilities of all voxel grids are used as weights, and the 3D coordinates of the voxel grids are summed together to obtain prediction results with smaller deviations.

\section{Experiment}
\subsection{Dataset and evaluation metrics}

\par We conducted experiments on three publicly available multi-view multi-person human pose estimation benchmarks: CMU-panoptic, Shelf, and Campus.
\par \textbf{CMU Panoptic.} CMU Panoptic provides some examples with large social interaction. It uses 480 synchronized VGA cameras, 31 synchronized HD cameras (temporally aligned with VGA cameras), and 10 RGB-D sensors for motion capture. All of the 521 cameras are calibrated by the Structure From Motion approach. Following\cite{tu2020voxelpose}, the training set consists of the following sequences: ``60422 ultimatum1",``16022 4 haggling1",``160226 haggling1",``161202 haggling1",``160906 ian1",``160906
ian2",``160906 ian3",``160906 band1",``160906 band2",``160906 band3". The testing set consists of :``160906 pizza1",``160422 haggling1",``160906 ian5",and ``160906 band4".

\par \textbf{Shelf and Campus.} The Shelf dataset has annotated the body joints of four actors interacting with each other using cameras 2, 3 and 4. Triangulation is performed using the three camera views for deriving the 3D ground truth. Actor 4 (Vasilis) is occluded in most of the camera views and thus excluded from the evaluation. The Campus dataset has annotated the body joints of the main three actors performing different actions for the frames that are observed from the first two cameras. The ground-truth for the third camera view is the result of the triangulation (between cameras 1 and 2), and then projected to camera 3.

\par We have used the following evaluation metrics on these datasets.
\par \textbf{PCP3D}(Percentage of Correctly estimated Parts)The PCP metric measures the percentage of correctly predicted parts. A body part is considered correct by the algorithm if:

\begin{equation}
\frac{\left.|s_n-\widehat{s_n}\right.|+\left.|e_n-\widehat{e_n}\right.|}{2}\le\alpha\left.|s_n-e_n\right.|
\end{equation}
where $s_n$ and $e_n$ are the ground truth start and end location of part $n$, $\widehat{s_n}$ and $\widehat{e_n}$ are the corresponding estimated locations, and $\alpha$  is a threshold parameter.

\par \textbf{MPJPE}(Mean Per Joint Position Error) This metric is calculated by:

\begin{equation}
E_{MPJPE}\left(f,S\right)=\frac{1}{N_s}\sum_{i=1}^{N_s}{\left.|P_{f,S}^{\left(f\right)}\left(i\right)-P_{gt,S}^{\left(f\right)}\left(i\right)\right.|_2}
\end{equation}
where f denotes a frame and s denote the corresponding skeleton. $P_{f,S}^{\left(f\right)}\left(i\right)$ is the estimated position of joint I and $P_{gt,S}^{\left(f\right)}\left(i\right)$ is the corresponding ground truth position. All joints are considered, $N_s$ means the number of joints. Finally, the MPJPEs are averaged over all frames.

\subsection{Implementation Details}

\par We test the performance on these three datasets under two scenarios, one is using the CPN network to predict the human center, the other is using the ground truth of the human center. The pretrained 2d human pose network backbone is used in all three datasets and require gradient is set to false. The embedding size of the Sinkhorn Transformer and the channels that 3D Conv extract are both set to 256. We set the depth (number of attention layers) to 1, size of the bin to 128, and number of attention heads for the Sinkhorn Transformer to 2. The input grid size is $32 \times 32 \times 32 $. We trained our model on a Nvidia GTX 3090 GPU.
\par For the Campus and Shelf datasets, we also adopt the same approach as \cite{tu2020voxelpose}, directly using the 2D pose estimator trained on the COCO dataset. These two datasets are only used as test sets, and the training process only uses synthetic data from the CMU panoptic dataset. We apply the Adam optimizer and train the model for 30 epochs. The learning rate is set to be 1e-4. The PCP3D metric is used to evaluate the predicted poses. For each ground truth pose, the most similar predicted pose is found and the percentage of correct limbs is calculated. However, we argue that this metric does not penalize false positive predictions, and does not provide an accurate analysis of errors in terms of distance. Therefore, we also measure the MPJPE (Mean Per Joint Position Error).

\par For the CMU Panoptic dataset, it is partitioned into a training set and a test set as described in the previous section, we apply the Adam optimizer and train the model for 10 epochs. The learning rate is also set to be 1e-4. The $AP_K$ metric in \cite{tu2020voxelpose} is used for evaluation on this dataset to facilitate comparison of performance. A prediction was considered to be estimated correctly when its MPJPE was less than $K$ mm.

\subsection{Experimental results}

\begin{figure*}[!htb]
 
\includegraphics[width=\linewidth]{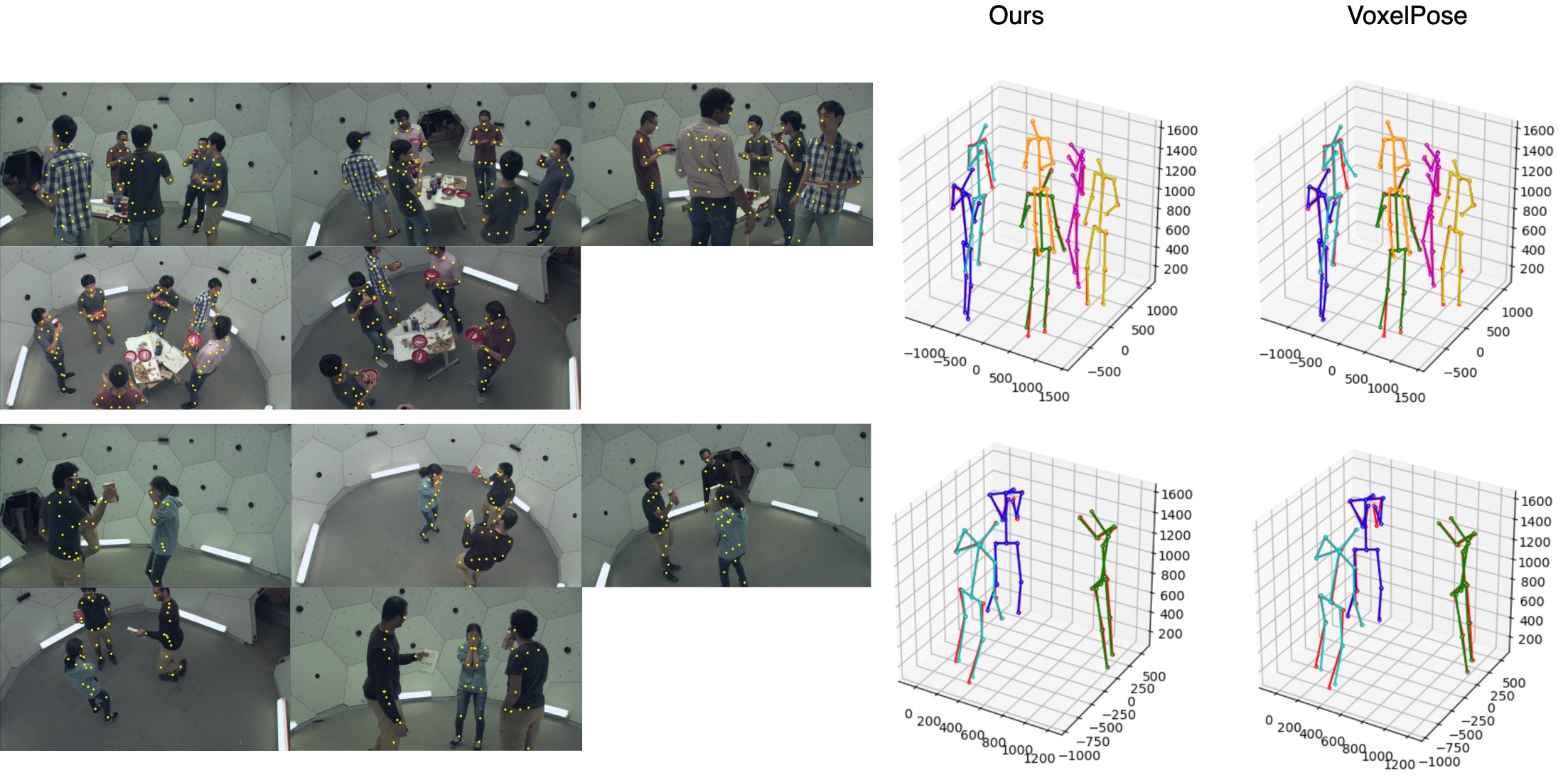}
  \caption{Estimated 3D poses using ground truth of human center under CMU Panoptic dataset}
  \label{fig:result_panoptic}
\end{figure*}

\par \textbf{Shelf and Campus.}  The two datasets are evaluated using the PCP3D metric for actor 1 (A1), actor 2 (A2), and actor 3 (A3), respectively, and the MPJPE results are also provided. Table \ref{tab:tab_shelf} presents the result for the Shelf dataset and Table \ref{tab:tab_campus} presents the result for the Campus dataset.

\begin{table}[!htbp]
\centering
\caption{Results on Shelf. $(t)$ means method with temporal information. $(gt)$ means using the ground truth body centers in people localization. $\uparrow$ means the larger the better, and $\downarrow$ means the lower the better.}

\resizebox{\linewidth}{!} {
\begin{tabular}{cccccc}
\hline
Shelf & A1 $\uparrow$& A2 $\uparrow$& A3 $\uparrow$& Avg. $\uparrow$& MPJPE (mm) $\downarrow$\\
\hline
$(t)$ 4D Association \cite{zhang20204d} & 99.0 & 96.2 & 97.6 & \textbf{97.6} & -\\
$(t)$ Part-aware Pose \cite{chu2021part} & 99.14 & 95.41 & 97.64 & 97.39 & -\\
$(t)$ VoxelTrack \cite{zhang2021voxeltrack} & 98.6 & 94.9 & 97.7 & 97.1 & -\\

\hline

Wu et al. \cite{wu2021graph} & 99.3 & 96.5 & 97.3 & \textbf{97.7} & - \\

Dong et al. \cite{dong2019fast} & 98.8 & 94.1 & 97.8 & 96.9 & - \\
VoxelPose \cite{tu2020voxelpose} & 99.3& 94.1 & 97.6 & 97.0 & 57.3\\
Ours & 99.3 & 95.1 & 97.4 & 97.3 & \textbf{56.3} \\
\hline
$(gt)$ VoxelPose \cite{tu2020voxelpose} & 99.2 & 95.1 & 97.8 & 97.4 & 52.8\\
$(gt)$ Ours  & 99.5 & 96.2 & 97.6 & \textbf{97.8} & \textbf{51.1}  \\
\hline
\end{tabular}
}
\label{tab:tab_shelf}
\end{table}

\par In the Shelf dataset, we surpassed VoxelPose in both PCP3D average performance and MPJPE, achieving PCP3D average performance of 97.3 and MPJPE of 56.3mm, respectively, compared to VoxelPose's PCP3D average performance of 97.0 and MPJPE of 57.3mm. When the ground truth of the body center is used in the first stage, our lead in PCP3D average performance comes to 0.4\%, and MPJPE reduces by 1.7mm, showing more significant improvement than under the condition of using CPN to predict the human center. This indicates that accurate localization of the targets can lead to highly accurate pose estimation results. 

\begin{table}[!htbp]
\centering
\caption{Results on Campus. $\uparrow$ means the larger the better, $\downarrow$ means the lower the better.}
\resizebox{\linewidth}{!} {
\begin{tabular}{cccccc}
\hline
Campus & A1 $\uparrow$ & A2 $\uparrow$ & A3 $\uparrow$ & Avg. $\uparrow$ & MPJPE (mm) $\downarrow$\\
\hline
$(t)$ Part-aware Pose  \cite{chu2021part} & 98.37 & 93.76 & 98.26 & \textbf{96.79} & -\\
$(t)$ VoxelTrack \cite{zhang2021voxeltrack} & 98.1 & 93.7 & 98.3 & 96.7 & -\\
\hline
Dong et al. \cite{dong2019fast} &  97.6 & 93.3 & 98.0 & 96.3 & - \\
VoxelPose \cite{tu2020voxelpose} & 97.6 & 93.8 & 98.8 & \textbf{96.7} & \textbf{78.2}\\
Ours & 97.6 & 93.1 & 98.1 & 96.3 & 80.1\\

\hline
$(gt)$ VoxelPose \cite{tu2020voxelpose}  & 97.6 & 93.3 & 98.7 & \textbf{96.5}  & 73.9 \\
$(gt)$ Ours  & 97.8 & 93.8 & 97.3 & 96.3 & \textbf{71.3} \\
\hline
\end{tabular}
}
\label{tab:tab_campus}
\end{table}

\par For the Campus dataset, our results in terms of $AP$ is on par with the state-of-the-art methods including VoxelPose. The MPJPE of our method is 2mm smaller than VoxelPose when using the ground truth in the center of the body, reaching 71.3mm compared with 73.9mm from VoxelPose. It should be noted that since VoxelPose did not test the performance when using the ground truth of human center, the results here are obtained by us using the official code, after 30 epochs of training, we get the result PCP3D average performance of 96.5. We speculate that it might be because of the smaller number of views in the Campus dataset, which only has 3 cameras (one camera's ground truth is the result of triangulation from the other two cameras), compared with 5 cameras from the other two datasets.

\par \textbf{CMU Panoptic.} We compared our proposed method VTP and VoxelPose under the Panoptic dataset. As shown in Table \ref{tab:tab_CMU}, our network improves by 1.28 mm compared to VoxelPose when using the ground truth of the body center. When using the CPN network as the first stage to predict the body center, our network improves 0.06mm in MPJPE compared to the already very accurate voxelpose, and our model has a higher correct rate under the threshold of $AP_{25}$.

\begin{table}[ht]
\centering
\caption{Results on CMU Panoptic.}
\resizebox{\linewidth}{!} {
\begin{tabular}{cccccc}
\hline
CMU Panoptic & $AP_{25}\uparrow$ & $AP_{50}\uparrow$ & $AP_{100}\uparrow$ & $AP_{150}\uparrow$ & MPJPE (mm) $\downarrow$\\
\hline
VoxelTrack \cite{zhang2021voxeltrack} & 85.88 & 98.31 & 99.54 & - &  16.97\\
\hline
Wu et al. \cite{wu2021graph} & - & - & - & - & \textbf{15.84} \\
VoxelPose \cite{tu2020voxelpose} & 83.59 & 98.33 & 99.76 & 99.91 & 17.68\\
Ours & 83.79 & 97.14 & 98.15 & 98.40 &  17.62 \\
\hline
$(gt)$ VoxelPose \cite{tu2020voxelpose} & - & - & - & - & 16.94\\
$(gt)$ Ours & - & - & - & - & \textbf{15.66} \\
\hline
\end{tabular}
}
\label{tab:tab_CMU}
\end{table}

\begin{table*}[!h]
\centering
\caption{Ablation study on the Shelf dataset.}
\begin{tabular}{cccccccccc}
\hline
& Structure & Attention head & Embedding size & Grid size & A1 & A2 & A3 & Avg. & MPJPE (mm)\\
\hline
(a) & CNN  & - & 256 & ${32\times 32\times 32}$ & 97.3 & 83.7 & 97.5 & 92.9 & 64.3mm\\

(b) &  Transformer  & 2 & 256 & ${32\times 32\times 32}$ & 98.2 & 91.8 & 97.5 & 95.8 & 61.3mm\\

(c) &  CNN + Transformer  & 2 & 32 & ${32\times 32\times 32}$ & 97.6 & 89.7 & 97.0 & 94.8 & 68.1mm\\

% CNN + Transformer & 2 & 64 & ${32\times 32\times 32}$ & 97.9 & 93.7 & 97.2 & 96.3 & 64.4mm\\

% CNN + Transformer & 2 & 128 & ${32\times 32\times 32}$ & 99.2 & 94.3 & 97.5 & 97.0 & 57.5mm\\

(d) &  CNN + Transformer & 2 & 256 & ${32\times 32\times 32}$ & 99.5 & 94.9 & 97.6 & 97.3 & 56.3mm\\

(e) &  CNN + Transformer & 4 & 256 & ${32\times 32\times 32}$ & 99.2 & 94.3 & 97.5 & 97.0 & 57.2mm\\

(f) &  CNN + Transformer & 8 & 256 & ${32\times 32\times 32}$ & 99.3 & 93.0 & 97.6 & 96.6 & 57.1mm\\

(g) &  CNN + Transformer & 2 & 512 & ${32\times 32\times 32}$ & 99.4 & 93.2 & 97.5 & 96.7 & 56.0mm\\

(h) &  CNN + Transformer(2 layers) & 2 & 256 & ${32\times 32\times 32}$ & 98.8 & 94.5 & 97.6 & 97.0 & 56.4mm\\

(i) &  CNN + Transformer & 2 & 256 & ${24 \times 24\times 24}$ & 99.2 & 94.3 & 97.7 & 97.0 & 58.1mm\\
\hline
\end{tabular}
\label{tab:ablation}
\end{table*}

\begin{figure*}[!htp]
  \includegraphics[width=\linewidth]{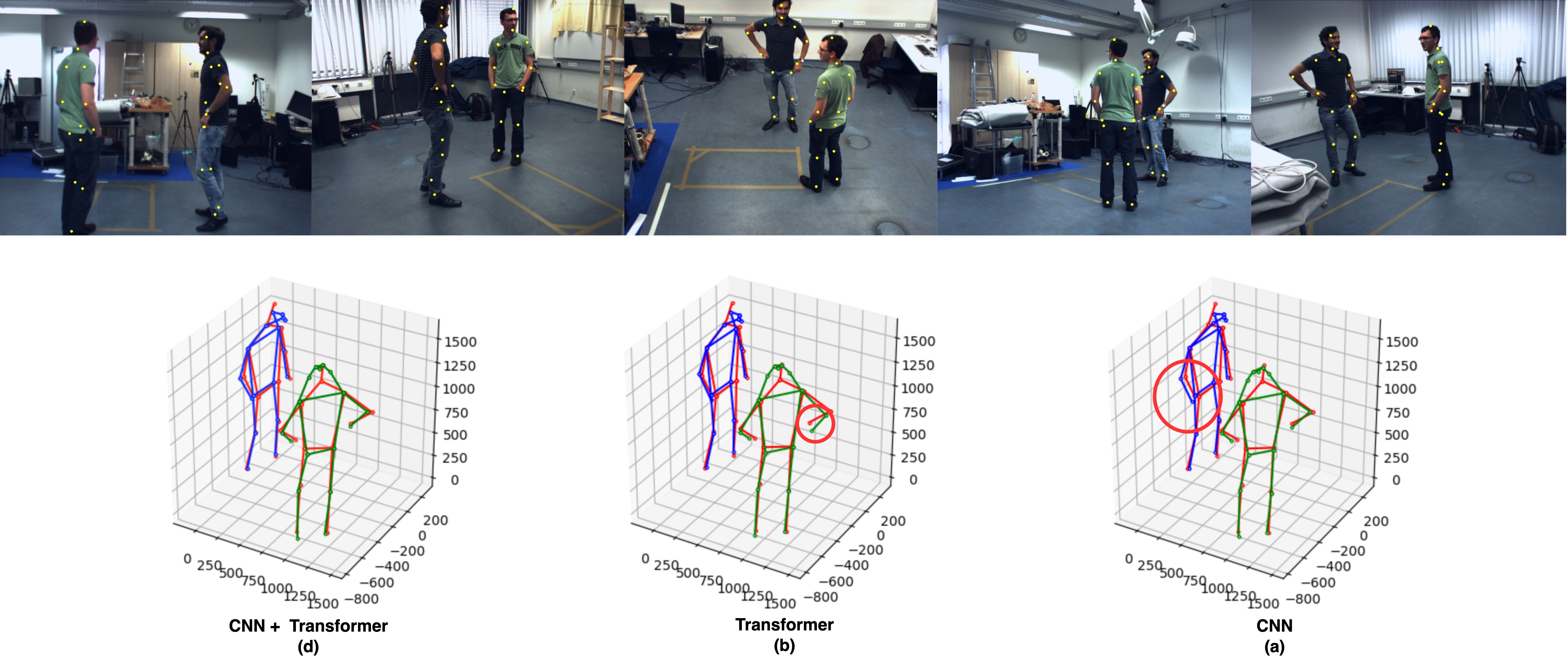}
  \caption{result on Shelf with different branch structure}
  \label{two branch}
\end{figure*}

\par Here we can find that after using the CPN network for prediction, our lead for voxelpose shrinks, and the same happens for the other two datasets. Our speculation is that because the body center predicted by the CPN network has some error compared to the ground truth, the predicted value may be randomly distributed in the vicinity of the ground truth, resulting in ambiquity in the relative relationship between the learned embedding of the transformer for voxel grid in the expression space. To this end, more accurate body center estimation is still needed.
Figure \ref{fig:result_panoptic} shows the visualization of the pose prediction using the ground truth of the human center in two different scenes. On the left, the original images with labeled 2D joints are shown in five views; on the right, the ground-truth poses of each person are represented by various colors, and the red color represents the predicted poses.
% In the first example, we can see that VoxelPose has more obvious errors in the prediction of the person in aqua color for the left arm and the person in green color for the left ankle, while our method has relatively small errors in these two parts. In the second example, the same happens for the left arm of the person in blue color, and both feet of the person in aqua color.

\subsection{Ablation studies}

\begin{figure*}[!htp]
  \includegraphics[width=\linewidth]{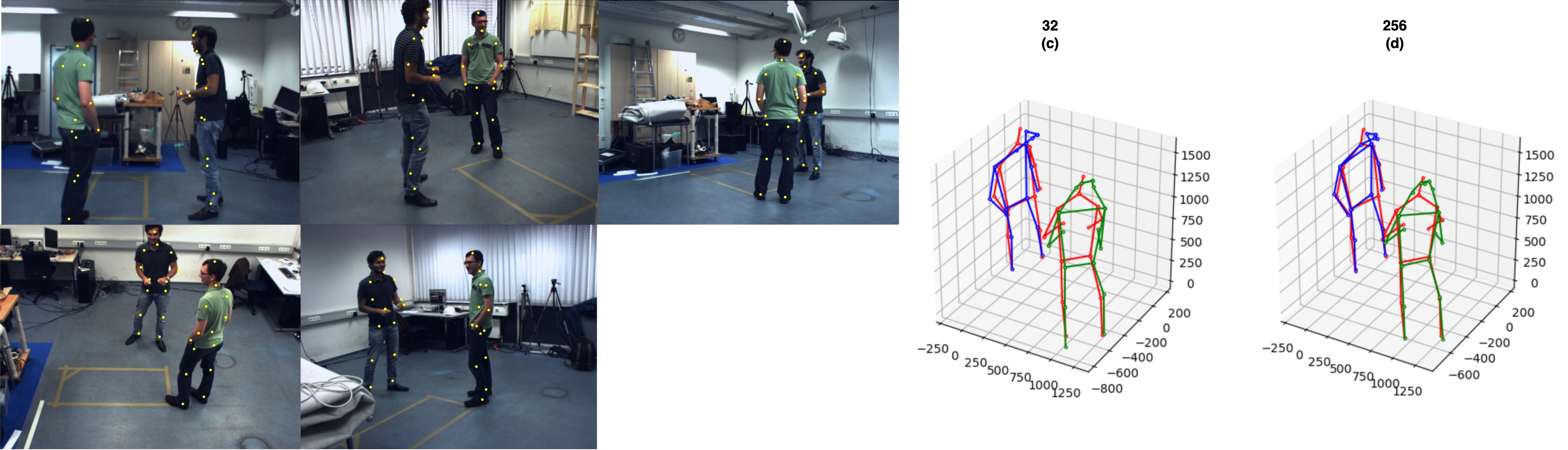}
  \caption{Results on Shelf with different embedding size.}
  \label{embedding result}
\end{figure*}

\par In this section,ablative studies are conducted to analyze the essential components in our proposed model in detail. Various parameters in our proposed VTP model are tested on the Shelf dataset, and the results are shown in Table \ref{tab:ablation}.

\par \textbf{Two branch structure}.
Our proposed network can be divided into three parts,
a separate convolution-based network(marked as (2) in Figure \ref{overview}) to increase the number of channels, a transformer-based network(including (1) in Figure \ref{overview} and following transformer encoder), the Embedding phase in Transformer can be regarded as an increase to the number of channels, finally is the part that reduces the number of channels from the above two parts jointed features to the number of joint nodes through convolutional layer(marked as (3) in Figure \ref{overview}). (a) and (b) of Table \ref{tab:ablation} show the effect of using only CNN to increase the number of channels and using only the transformer network to increase the number of channels. It can be found that the effect of using these two branches alone is much worse than that of using the features of both branches in (d). Figure \ref{two branch} shows a result using different branches.
\par The limited receptive field of CNN leads to difficulty in capturing global information, while Transformer can capture long-range dependencies, so combining the two enables the network structure to inherit the advantages of CNN and Transformer and retain the global and local features to the greater extent.

\par
\begin{figure}[htp]
  \includegraphics[width=\linewidth]{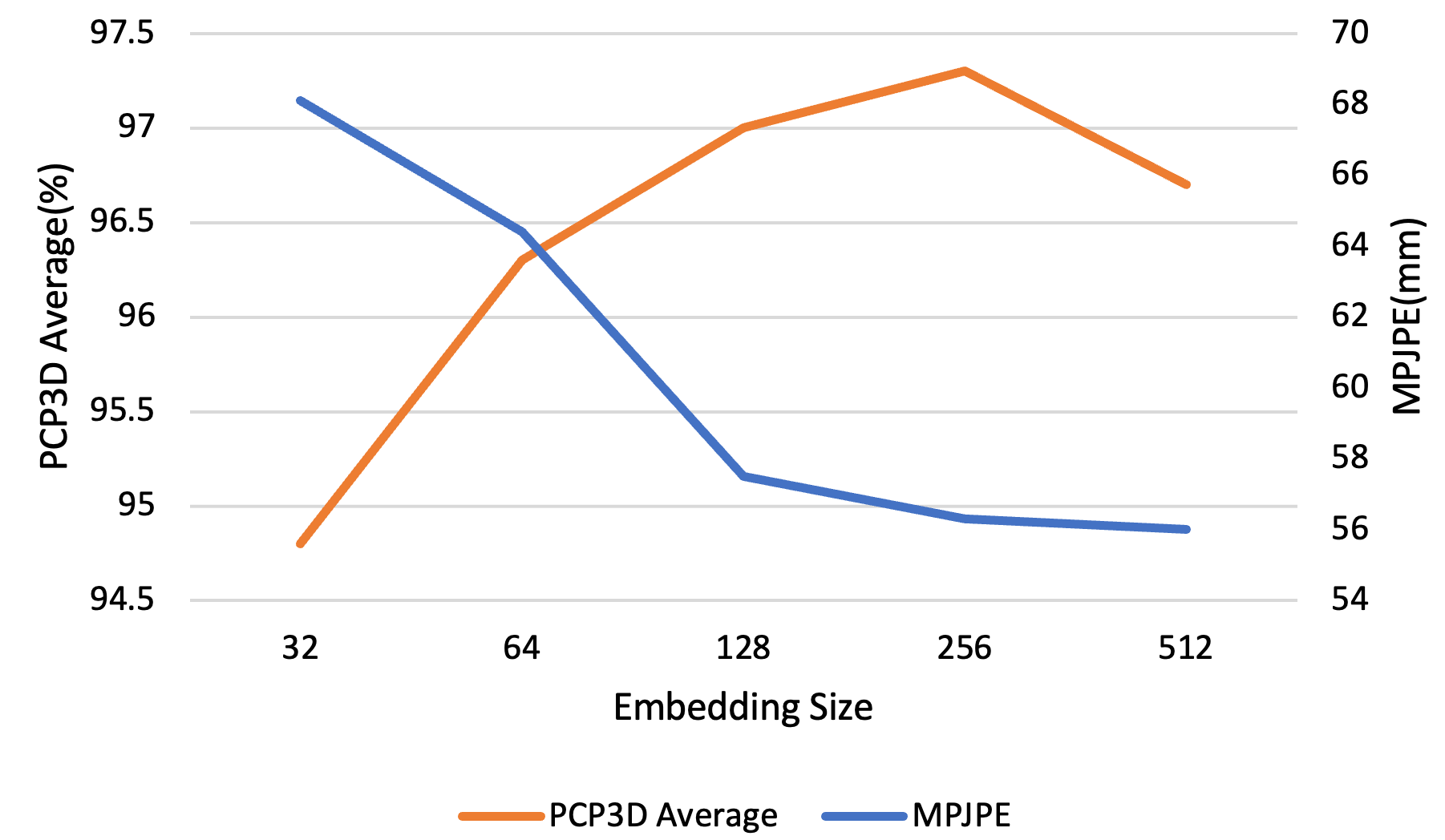}
  \caption{Relation between embedding size and PCP3D Average / MPJPE.}
  \label{fig:embedding}
\end{figure}

 \textbf{Embedding size}.
We study the effect of choosing the embedding size. As shown in the Table \ref{tab:ablation} (c), (d), (g), we change the embedding size of the transformer branch and the number of output channels of the CNN branch, while all other parameters are fixed. The MPJPE increases to 68.1mm and the PCP3D average performance is only 94.8 when the embedding size is set to 32. The performance in terms of these two metrics is improving as the embedding size increases from 32 to 256. However, when the embedding size increases from 256 to 512, the improvement of MPJPE is not obvious, also the PCP3D Average performance decreases by 0.6\%. This may be owing to that the 512 embedding size made the model too complex, overfitting the training set, thus deteriorating the generalization ability. Figure \ref{fig:embedding} shows the performances for five different embedding sizes. The performance improvement is more obvious in the process of increasing the embedding size from 32 to 128, and the effect starts to slow down from 128 to 256. Figure \ref{embedding result} shows a result using 32 and 256 embedding sizes.

\par \textbf{Number of layers and attention heads.} One of the major bottlenecks of using transformers on voxel grids is the complexity, where the increase in granularity of the 3D space will bring a cubic increase in the length of the transformer sequence. Increasing the number of layers will consume more memory resources linearly, unless the reversible layers like the one in reformer\cite{kitaev2020reformer} are used, which saves space to store the intermediate gradients but significantly reduce the training speed. In Table \ref{tab:ablation}(d) and (h), we compare the results of 1 transformer layer versus 2 transformer layers. In (h) we train 40 epochs in case the network does not converge. The MPJPE decreases slightly after increasing the transformer to two layers, and the PCP3D average performance decreases by 0.3. Increasing the number of layers to 2 behaves similarly to increasing the embedding size to 512.

\par For the multi-headed attention mechanism, we did not find any obvious pattern. As shown in Table \ref{tab:ablation}(d-f), we test the effects under 2, 4, and 8 attention heads, respectively. The MPJPE and PCP3D average performance of 2 heads are the highest, achieving 56.3mm and 97.3 respectively, while under 4 and 8 heads, the PCP3D average decreases continuously, while the MPJPE decreases and then increases. Therefore, we choose the 2 head attention mechanism for our final architecture.

\par \textbf{Grid size}. By comparing (d),(i) in Table \ref{tab:ablation},  we can see that the performance of both MPJPE and PCP3D is improved when the grid size (in other words, the number of grids that the space is divided into) is increased from ${24\times 24\times 24}$ to ${32\times 32\times 32}$. It shows that a more detailed division of space brings better results, which always holds true for voxel-based methods. However, as mentioned before, a more detailed division of the space will lead to a significant increase in computation. We choose ${32\times 32\times 32}$ in order to strike the balance between complexity and precision.

\section{Conclusion and Future Work}
\par In this paper, we propose the Volumetric Transformer Pose estimator (VTP), a network consisting of transformer and CNNs, for volumetric representation learning in multi-view multi-person 3D pose estimation scenarios. In the two-stage process of localizing multiple people and estimating the pose for each person, VTP contributes to better volumetric representation learning in the second stage, where the pose of the localized person is estimated. A residual structure and the Sinkhorn attention are applied to further improve the efficiency and accuracy. Experiments on popular benchmarks show that VTP is on par with the state-of-the-art methods in terms of PCP3D and MPJPE, and outperforms previous voxel-based methods when the target person localization is accurate. In future works, we will explore the potential of the proposed framework in the first stage of localizing the persons as well, which will likely increase the overall accuracy.

\printbibliography


% \end{IEEEbiography}

\vfill

\end{document}